# Control Architecture and experimental validation of a Novel Surgical Robotic Instrument

1st Doina Pisla
*CESTER, Dept. of Mechanical Engineering*
*Technical University of Cluj-Napoca*
*European University of Technology, EU*
*Technical Sciences Academy of Romania, Bucharest*
Cluj-Napoca, Romania
doina.pisla@mep.utcluj.ro

2nd Ionut Zima*
*CESTER, Dept. of Mechanical Engineering*
*Technical University of Cluj-Napoca*
*European University of Technology, EU*
Cluj-Napoca, Romania
ionut.zima@mep.utcluj.ro

3rd Calin Vaida
*CESTER, Dept. of Mechanical Engineering*
*Technical University of Cluj-Napoca*
*European University of Technology, EU*
Cluj-Napoca, Romania
calin.vaida@mep.utcluj.ro

4th Andrei Cailean
*CESTER, Dept. of Mechanical Engineering*
*Technical University of Cluj-Napoca*
*European University of Technology, EU*
Cluj-Napoca, Romania
andrei.cailean@mep.utcluj.ro

5th Marius Miclaus
*CESTER, Dept. of Mechanical Engineering*
*Technical University of Cluj-Napoca*
*European University of Technology, EU*
Cluj-Napoca, Romania
Marius.Miclaus@mep.utcluj.ro

6th Adrian Pisla
*CESTER, Dept. of Mechanical Engineering*
*Technical University of Cluj-Napoca*
*European University of Technology, EU*
Cluj-Napoca, Romania
adrian.pisla@muri.utcluj.ro

7th Andrei Caprariu
*CESTER, Dept. of Mechanical Engineering*
*Technical University of Cluj-Napoca*
*European University of Technology, EU*
Cluj-Napoca, Romania
Andrei.Caprariu@mep.utcluj.ro

8th Vasile Bulbucan
*CESTER, Dept. of Mechanical Engineering*
*Technical University of Cluj-Napoca*
*European University of Technology, EU*
Cluj-Napoca, Romania
vasile.bulbucan@mep.utcluj.ro

9th Bogdan Gherman
*CESTER, Dept. of Mechanical Engineering*
*Technical University of Cluj-Napoca*
*European University of Technology, EU*
Cluj-Napoca, Romania
bogdan.gherman@mep.utcluj.ro

10th Damien Chablat
*CESTER, Dept. of Mechanical Engineering*
*Technical University of Cluj-Napoca*
*European University of Technology, EU*
*LS2N, UMR 6004, CNRS*
*École Centrale Nantes, Nantes Université*
Nantes, France
damien.chablat@cnrs.fr

*Abstract* — **Minimally invasive surgery (MIS) reduces patient trauma and shortens recovery time; however, conventional laparoscopic instruments remain constrained by limited range of movements. This work presents the control architecture of a 4-DOF flexible laparoscopic instrument integrating distal bending, independent distal head rotation, shaft rotation, and a gripper, while maintaining a 10 mm diameter compatible with standard trocars. The actuation unit and SpaceMouse teleoperation are implemented on Raspberry Pi 5 with Motoron controllers. An analytical scissor-linkage model is derived and parameterized. The predicted jaw opening corresponds to CAD measurements (MAE 0.13°) and OptiTrack motion capture (MAE 1.43°). Integration with the ATHENA parallel robot is validated through a simulated pancreatic surgery procedure.**

## I. INTRODUCTION

Minimally invasive surgery (MIS) revolutionized surgical procedures since the first laparoscopic cholecystectomy in 1988 [1]. The main advantages of robotic procedures such as MIS are reduced postoperative pain, shorter hospitalization, faster recovery, and decreased blood loss [2-6]. However, MIS presents technical challenges such as: restricted degrees of freedom [7], fulcrum effect [8], loss of haptic feedback [9], limited depth perception [10], and tremor amplification [3]. Robot-Assisted MIS (RAMIS) addresses these limitations via improved visualization, motion scaling, and tremor filtering [11]. The da Vinci Surgical System (Intuitive Surgical, Inc., USA) with EndoWrist (Intuitive Surgical, Inc., USA) instruments provides wrist-like articulation improving surgical dexterity [12-15]. An important challenge is represented by the trade-off between flexibility and stiffness [16,17].

Several flexible instruments have been developed: Kanno et al. [18] presented a 4-DOF forceps manipulator; Haraguchi et al. [19] integrated force sensing; Hong and Jo [20] focused on improved rigidity; Gherman et al. [21] developed a flexible design successfully integrated with the ATHENA robot [22,23]; Schmitz et al. [24] proposed a rolling-tip mechanism; Hong et al. [17] presented hysteresis compensation; Wang et al. [25] developed a hybrid snake-like robot and Wu et al. [26] presented a modular continuum instrument.

This paper presents the control and experimental validation of a 4-DOF flexible laparoscopic instrument with a 10 mm diameter. Contributions include: (1) analytical modeling of the scissor-linkage gripper mechanism; (2) control architecture for teleoperation of the 4-DOF instrument using Raspberry Pi 5 (Raspberry Pi Ltd., UK) with

SpaceMouse input (3Dconnexion, Germany); (3) experimental validation using OptiTrack (NaturalPoint, Inc., USA) motion capture [23] and integration with the Athena robot. The paper is organized as follows: Section II describes mechanical architecture; Section III presents kinematic modeling; Section IV details the control system; Section V presents experimental validation including the integration of the instrument on the Athena robot and Section VI presents the conclusions.

## II. MECHANICAL STRUCTURE OF THE INSTRUMENT

The proposed flexible laparoscopic instrument [27,28] has a 10 mm outer diameter, compatible with standard laparoscopic trocars. The four degrees of freedom (DOF) of the proposed instrument are: bending of the flexible element ($q_1$), rotation of the distal head ($q_2$), gripper actuation ($q_3$) and rotation of the entire rod ($q_4$). The architecture of the instrument is illustrated in Fig. 1. The four main sections of the instrument are: the actuation unit, the rigid rod, the flexible element, and the distal head. The actuation unit incorporates motors and electronics. The rigid rod transmits actuation forces from the driving unit to the flexible element and enables rotation about the longitudinal axis ($q_4$). The flexible element provides controlled bending using a series of triangular notches. The distal head contains the gripper mechanism and enables independent rotation ($q_2$) while the instrument is bent.

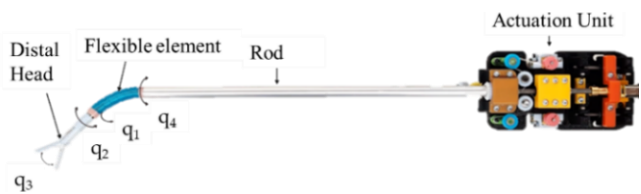

Fig 1. Overview of the flexible laparoscopic instrument

The flexible element is 3D printed as a monolithic structure using Elastico PolyJet Photopolymer (Stratasys Ltd., USA) on a Stratasys MediJet 3D printer [23]. This design combines the elastic joint with rigid segments and integrated wire guidance channels. The flexible section of the instrument presents a series of triangular notches distributed along the tube, as can be observed in Fig. 2. These notches function as compliant hinges, enabling controlled bending motion ($q_1$) in a single plane. Each notch is characterized by a half-angle $\alpha$ = 7.5°, with the full notch angle ($2\alpha$ = 15°). The design uses $n$ = 6 notches per side, achieving a maximum total bending angle of $\theta_{max}$ = 90°.

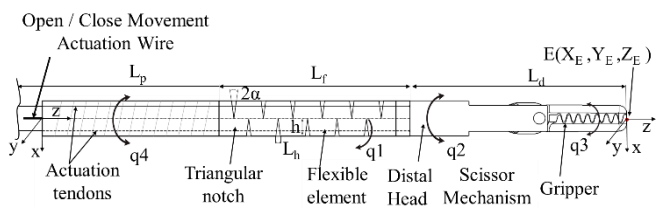

Fig 2. Flexible element and Distal Head

When the bending tendons are actuated, notches on the compression side progressively close while notches on the tension side open. Rigid tubes are embedded in the flexible element structure to guide the bending and unbending wires while another channel in the center of the flexible element is used for the gripper actuation wire. Another set of channels is used for the Nitinol reinforcement segments that are inserted to improve stiffness of the flexible segment.

Two antagonistic pairs of actuation tendons control the bending and unbending movements. Pulling the flexion wires while releasing the extension wires produces bending while reverse movements for the unbending.

The distal head incorporates a scissor linkage mechanism that enables gripper actuation ($q_3$). The grasping function is implemented using a bidirectional push-pull wire mechanism: when the wire is pushed, the gripper jaws open and when pulled, the jaws close.

An important feature of the instrument is represented by the independent rotation of the distal head ($q_2$) through 360° via a flexible metallic wire transmission as illustrated in Fig.1 and Fig.2. This capability allows grasper reorientation while the instrument remains in a bent configuration.

## III. KINEMATIC AND FORCE MODELING

This section presents the mathematical model of the scissor-linkage gripper mechanism, including kinematic equations linking the slider displacement to the total jaw opening angle, and force propagation equations integrated in the control of the instrument for real-time position control. The mathematical modeling of the scissor mechanism follows the geometrical analysis described in [29].

### A. Scissor Linkage Geometry

A symmetric scissor-linkage mechanism that converts linear slider motion into angular jaw rotation. Due to symmetry, only one half of the mechanism is modeled.

Fig.3 presents (a) the scissor linkage grasper and the fixed pivot point $P$, (b) the full free-body diagram (FBD) highlighting the input shaft force $F_{IN}$ and the output jaw tip forces $F_T$, and (c) the half-model used for derivation by applying the line of symmetry.

The following assumptions were adopted during the analysis: (1) quasi-static operation; (2) rigid linkages with negligible mass; (3) negligible friction at joints; (4) symmetric load between both jaws.

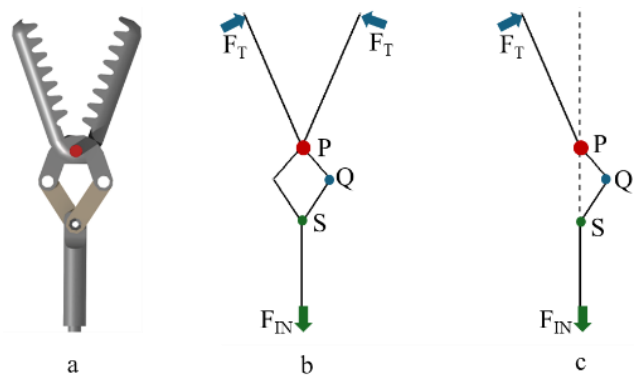

Fig 3. Laparoscopic gripper: (a) CAD model, (b) free-body diagram (FBD), and (c) half-model FBD

### B. Force Propagation Model

An input force $F_{IN}$ is applied by the actuator at the slider. Because the mechanism is symmetric, each half of the gripper carries half of the input force.

Fig.4 shows the half-grasper FBD, including (a) the complete force components and (b) the simplified diagram containing only contributing components and the angles used in the derivation.

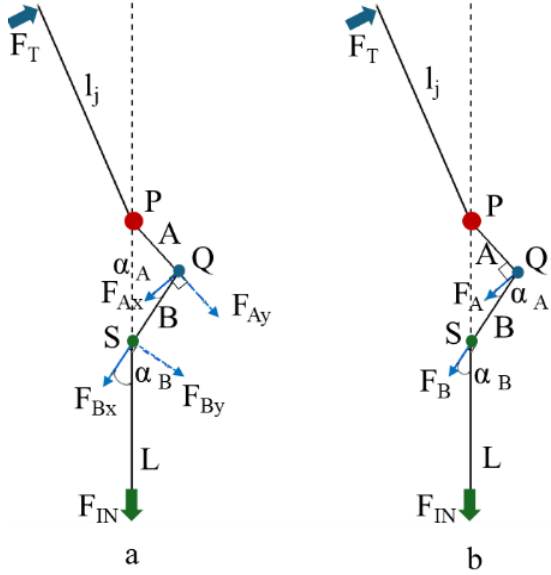

Fig 4. Half-grasper force analysis: (a) FBD, (b) simplified FBD

The half input force is:

$$F_S = \frac{1}{2} F_{IN} \qquad (1)$$

The force transmitted through linkage $B$ is the contributing component of $F_S$ projected along linkage $B$:

$$F_B = F_S \cos(\alpha_B) = \frac{1}{2} \cos(\alpha_B) F_{IN} \qquad (2)$$

At joint $Q$, the force is redirected through linkage $A$:

$$F_A = F_B \cos(\alpha_A) \qquad (3)$$

Taking moments about pivot $P$:

$$A F_A - l_j F_T = 0 \qquad (4)$$

The tip force per jaw is:

$$F_T = \frac{A}{l_j} F_A \qquad (5)$$

Substituting (2) and (3) into (5) yields:

$$F_T = \frac{A}{2 l_j} \cdot \cos(\alpha_A) \cdot \cos(\alpha_B) \cdot F_{IN} \qquad (6)$$

The combined grasping force applied by both jaws can be defined as:

$$F_{total} = 2 F_T = \frac{A}{l_j} \cdot \cos(\alpha_A) \cdot \cos(\alpha_B) \cdot F_{IN} \qquad (7)$$

C. Jaw Angle Model

The kinematic relationship between slider displacement $\Delta L$ and jaw opening angle is derived from the internal triangle formed by pivot $P$, joint $Q$, and slider point $S$. Fig.5. (a) illustrates the jaw angle definition and the jaw offset $\theta_O$, while Fig. 5 (b) presents the internal triangle parameters used to compute the angles required for both kinematic and force modeling.

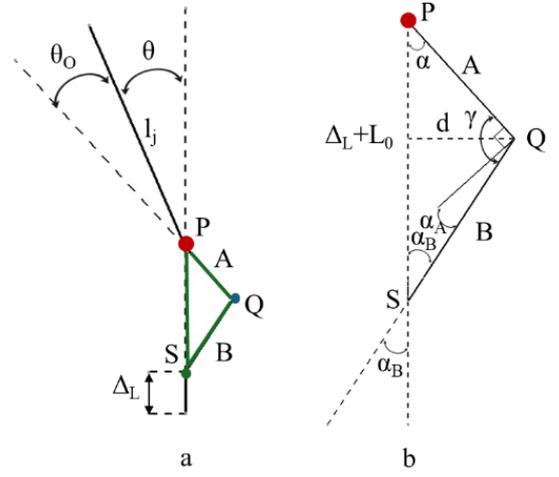

Fig 5. Kinematic geometry of the half-grasper: (a) Jaw offset angle, jaw angle, and linkage displacement, and (b) internal triangle parameters

The pivot-to-slider distance is:

$$PS = L_0 - \Delta L \qquad (8)$$

Applying the law of cosines to triangle PQS:

$$B^2 = A^2 + (L_0 - \Delta L)^2 - 2A(L_0 - \Delta L)\cos(\alpha) \qquad (9)$$

Solving for α:

$$\alpha = \arccos\left[\frac{B^2 - A^2 - (L_0 - \Delta L)^2}{-2A(L_0 - \Delta L)}\right] \qquad (10)$$

The jaw offset angle is:

$$\theta_O = \arcsin\left(\frac{h}{A}\right) \qquad (11)$$

The single jaw angle is:

$$\theta_{jaw} = \alpha - \theta_O \qquad (12)$$

The total angle between both jaws is:

$$\theta_{total} = 2\theta_{jaw} = 2(\alpha - \theta_O) \qquad (13)$$

The internal angles required for the force model are:

$$\alpha_B = \arcsin\left(\frac{A \sin(\alpha)}{B}\right) \qquad (14)$$

$$\alpha_A = \alpha + \alpha_B - 90° \qquad (15)$$

$$\alpha_A = \alpha + \arcsin\left(\frac{A \sin(\alpha)}{B}\right) - 90° \qquad (16)$$

The tip-to-tip opening width is:

$$W = 2 l_j \sin\left(\frac{\theta_{total}}{2}\right) \qquad (17)$$

TABLE I. GEOMETRIC PARAMETERS OF THE GRIPPER

| Parameter | Symbol | Value | Unit |
|---|---|---|---|
| Linkage A length | $A$ | 6.50 | mm |
| Linkage B length | $B$ | 8.00 | mm |
| Jaw length (pivot to tip) | $l_j$ | 22.3 | mm |
| Offset distance | $h$ | 2.5 | mm |
| Initial pivot-to-slider distance | $L_0$ | 13.6 | mm |
| Jaw offset angle | $\theta_0$ | 22.6 | deg |

Table 1 summarizes the geometric parameters of the mechanism.

## IV. CONTROL OF THE FLEXIBLE INSTRUMENT

This section presents the control architecture of the instrument as illustrate in the Fig.6.

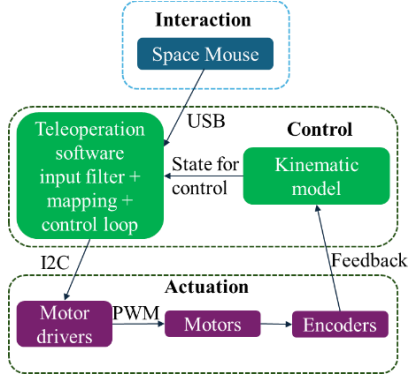

Fig 6. Control system architecture of the instrument

### A. Hardware Architecture

The central control unit is a Raspberry Pi 5 running the control software. Actuation is performed using two Motoron M2H18v18 (Pololu Corporation, USA) dual-channel motor controllers, providing independent control for the four degrees of freedom of the instrument. Each motor is equipped with a Pololu Romi Magnetic Encoder to provide feedback. The Raspberry Pi 5 operates at 5 V while the motor controllers are powered from a dedicated 12 V supply. The hardware architecture is illustrated in Fig. 7.

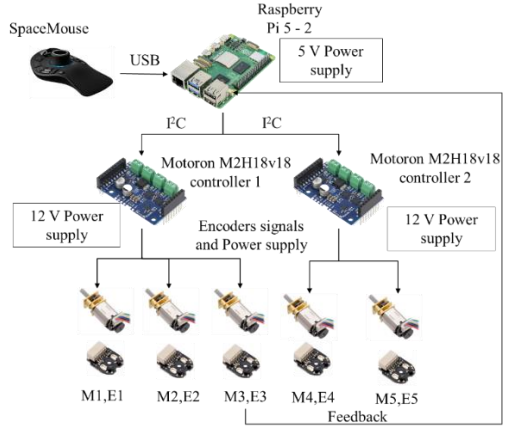

Fig 7. Hardware architecture of the flexible instrument

### B. System Block Diagram

The system architecture is organized into three layers: interaction, control, and actuation. The interaction layer comprises the surgeon input interface using a 3Dconnexion SpaceMouse, connected to the Raspberry Pi 5 via USB. The surgeon manipulates the SpaceMouse to control the movements of the instrument's distal end. The control layer is implemented on the Raspberry Pi 5 and consists of several processing stages. Raw SpaceMouse signals are acquired, normalized to the interval $[-1, 1]$, and passed through dead-zone and low-pass filters to eliminate hand tremors and micro-movements. The filtered values are then mapped and scaled into the required joint velocities for the four degrees of freedom ($q_1$- $q_4$). For bending movements, an antagonistic mapping converts the selected command into coordinated speeds for the flexion and extension motors, maintaining wire tension. The resulting velocity commands are transmitted via I²C to the Motoron drivers.

The kinematic model is integrated into the control layer to provide real-time state estimation. Encoder measurements from the gripper motor are converted to slider displacement $\Delta L$ using the transmission ratio. The current jaw opening angle θ is then computed.

The actuation layer contains the Motoron M2H18v18 motor drivers, the DC micro gear-motors, and the mechanical structure of the active instrument. The Motoron receive I²C commands from the Raspberry Pi and generate the corresponding PWM and direction signals for each motor

### C. Software Architecture

The control software runs on the Raspberry Pi 5 following a finite state machine structure with four states: INIT, IDLE, TELEOP, and FAULT, as illustrated in Fig. 8.

In the INIT state, the program configures the I²C bus, detects and initializes both Motoron M2H18v18 controllers, sets safe default limits for speed and acceleration, and opens the USB connection to the SpaceMouse. If all initialization checks succeed, the system transitions to IDLE; otherwise, it enters FAULT with all motors disabled.

In the IDLE state, the hardware is powered and responsive but no motion is commanded. Motor speeds are set to zero and the instrument maintains its current configuration. The software continuously monitors the SpaceMouse buttons; pressing the designated enable button transitions the system to TELEOP.

The TELEOP state contains the main periodic control loop executing at a fixed rate. At each cycle, the software: (1) reads the SpaceMouse axes and buttons, (2) applies dead-zone and low-pass filtering, (3) maps filtered values into joint velocity commands, (4) applies priority logic and antagonistic bending mapping, (5) reads encoder values and computes the current jaw angle using the kinematic model, and (6) sends updated speed commands to the Motoron controllers via I²C.

In the FAULT state, all motor commands are forced to zero. The controller remains in this state until a reset sequence is performed.

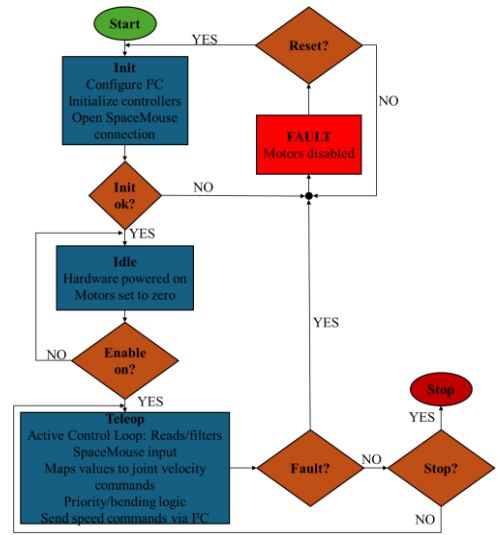

Fig 8. Software finite state machine diagram

### D. Master-Slave Configuration

The control scheme follows a classical velocity-based master-slave teleoperation configuration as illustrated in Fig.

9. The master side consists of the surgeon manipulating the SpaceMouse, while the slave side is the active laparoscopic instrument. The Raspberry Pi 5 acts as the coordinator between the two: it reads the master inputs, transforms them into joint-level velocity commands, and forwards the resulting speeds to the motor drivers.

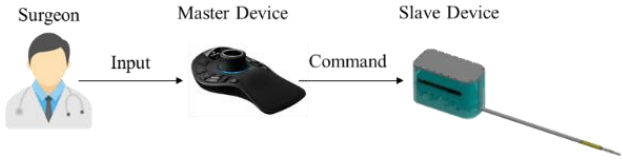

Fig 9. Master-slave configuration

Moving the SpaceMouse from its neutral position produces a proportional joint velocity; returning the knob to center stops the motion. This velocity-based control approach is intuitive and does not require explicit position synchronization between master and slave. The global feedback loop is closed visually by the surgeon through direct observation or endoscopic camera view.

*E. SpaceMouse Integration and Mapping*

The SpaceMouse provides six continuous analogue channels corresponding to three translational and three rotational motions of the knob, as well as several digital buttons. Each axis value is first normalized to the interval [-1,1] and passed through a small dead-zone to eliminate hand tremors and unintentional micro-movements.

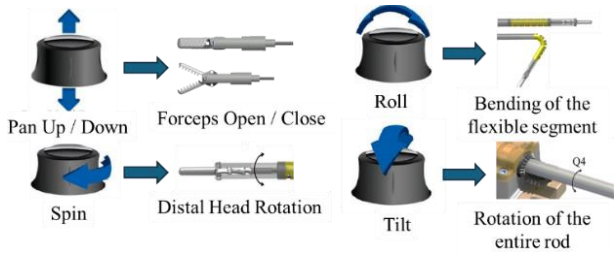

Fig 10. SpaceMouse axes mapping to instrument degrees of freedom

A simple low-pass filter further smooths the signals to avoid abrupt changes in motor speed. The filtered values are then mapped to the four degrees of freedom of the instrument as illustrated in Fig. 10. In the configuration used for this work, the vertical translation $T_z$ controls the opening and closing of the gripper $q_3$, the rotation $R_x$ around the x axis controls the rotation of the entire rod $q_4$, the rotation $R_y$ around the y axis controls the bending of the instrument $q_1$, and the axial rotation $R_z$ controls the rotation of the distal head $q_2$.

## V. EXPERIMENTAL EVALUATION

This section presents the experimental validation of the kinematic model and the integration of the flexible instrument with Athena robot. The analytical model is compared with CAD measurements and OptiTrack motion capture data [23].

*A. Kinematic Analysis Setup*

The kinematic model was validated using CAD measurements from nine configurations and experimental data acquired using an OptiTrack motion capture system. As illustrated in Fig. 11-12, passive reflective markers were placed on each jaw and on the moving flange to capture the jaw opening angle θ and slider displacement ΔL during an open-close cycle.

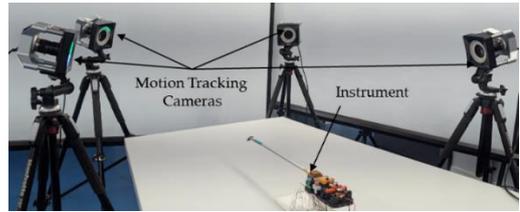

Fig 11. Motion analysis setup

For analysis, a full movement of the grasper was performed and the OptiTrack recorded the 3D coordinates of all markers. The jaw angle is computed as the angle between the two jaw vectors, and the displacement is computed from the distance between markers.

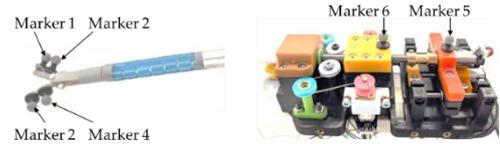

Fig 12. OptiTrack marker placement

*B. Kinematic Validation Results*

Fig. 13 illustrates the analytical model curve with both the CAD and OptiTrack measurement points.

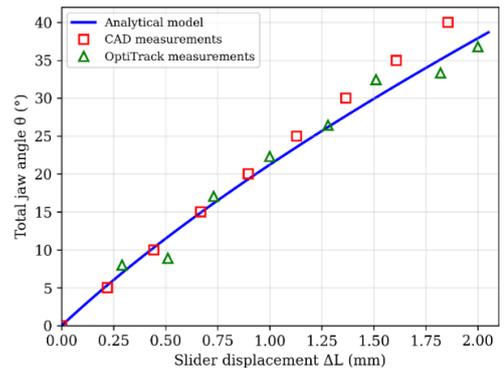

Fig 13. Kinematic validation

*C. Integration of the flexible instrument with Athena robot*

To demonstrate clinical applicability, the instrument was integrated with the ATHENA parallel robot for simulated pancreatic surgery. The setup illustrated in Fig. 14 includes the ATHENA robot, the flexible instrument, a laparoscopic camera, and 3D-printed organ phantoms.

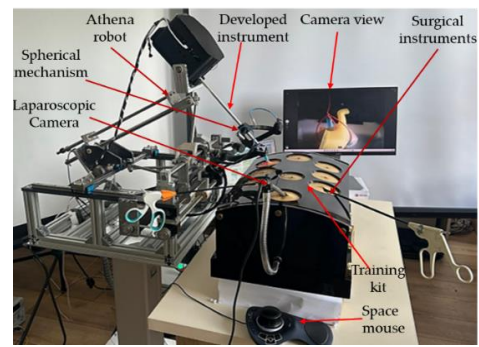

Fig 14. Integration of the surgical instrument with the Athena robot

*D. Discussion*

The experimental evaluation validates the analytical scissor-linkage model by comparing predicted jaw angles to independent measurements. Compared to CAD configurations, the model achieved MAE = 0.13°, confirming the geometric derivation. Compared to OptiTrack motion capture of the prototype, the MAE was 1.43°, the higher error is attributed to marker placement precision, out-of-plane motion, and clearances of the 3D printed joint of the instrument. Both results confirm that the model accurately predicts jaw kinematics within acceptable limits for surgical grasping applications. The control architecture running on the Raspberry Pi 5 enabled stable teleoperation of all four degrees of freedom during the experimental tests.

The successful integration with the ATHENA robot validated the instrument's capability to perform stomach retraction during a simulated pancreatic surgery procedure, demonstrating the feasibility of the proposed design for robot-assisted MIS.

## VI. Conclusions

This paper presents the control and experimental evaluation of a 4-DOF flexible laparoscopic instrument. The 10 mm diameter tool combines distal bending via a monolithic notched flexure, independent tip rotation, shaft rotation, and a scissor-linkage gripper. A SpaceMouse-based teleoperation framework was implemented. A key contribution is the analytical modelling of the scissor-linkage gripper. The model was validated using CAD measurements and the OptiTrack motion capture system.

The control architecture implemented on Raspberry Pi 5 with Motoron motor controllers demonstrated reliable coordination of the four degrees of freedom. The SpaceMouse-based teleoperation enabled intuitive master–slave control.

Experimental evaluation demonstrated successful integration with the ATHENA robot for pancreatic surgery simulation. Future work will focus on characterizing hysteresis, implementing closed-loop control and testing on tissue-representative phantoms.

## Acknowledgement

This work was funded by the project "Romanian Hub for Artificial Intelligence-HRIA", Smart Growth, Digitization and Financial Instruments Program, MySMIS no. 351416.